\documentclass[lettersize,journal]{IEEEtran}
\usepackage{amsmath,amsfonts}
\usepackage[linesnumbered,ruled,vlined]{algorithm2e}
\usepackage{array}
\usepackage[caption=false,font=normalsize,labelfont=sf,textfont=sf]{subfig}
\usepackage{textcomp}
\usepackage{stfloats}
\usepackage{url}
\usepackage{verbatim}
\usepackage{graphicx}
\usepackage{color}
\usepackage{hyperref}
\usepackage[numbers,sort&compress]{natbib}
\begin{document}

\title{Conformal Depression Prediction}

\author{Yonghong Li and Xiuzhuang Zhou

\IEEEcompsocitemizethanks{
\IEEEcompsocthanksitem This work was funded in part by the National Natural Science Foundation of China under Grants 61972046 (\textit{Corresponding author: Xiuzhuang Zhou}).
\IEEEcompsocthanksitem Yonghong Li and Xiuzhuang Zhou are with the School of Artificial Intelligence, Beijing University of Posts and Telecommunications, Beijing 100876, China  (Email: xiuzhuang.zhou@bupt.edu.cn).
}}




\maketitle

\begin{abstract}
While existing depression prediction methods based on deep learning show promise, their practical application is hindered by the lack of trustworthiness, as these deep models are often deployed as \textit{black box} models, leaving us uncertain on the confidence of their predictions. For high-risk clinical applications like depression prediction, uncertainty quantification is essential in decision-making. In this paper, we introduce conformal depression prediction (CDP), a depression prediction method with uncertainty quantification based on conformal prediction (CP), giving valid confidence intervals with theoretical coverage guarantees for the model predictions. CDP is a plug-and-play module that requires neither model retraining nor an assumption about the depression data distribution. As CDP provides only an average coverage guarantee across all inputs rather than per-input performance guarantee, we further propose CDP-ACC, an improved conformal prediction with approximate conditional coverage. CDP-ACC firstly estimates the prediction distribution through neighborhood relaxation, and then introduces a conformal score function by constructing nested sequences, so as to provide a tighter prediction interval adaptive to specific input. We empirically demonstrate the application of CDP in uncertainty-aware facial depression prediction, as well as the effectiveness and superiority of CDP-ACC on the AVEC 2013 and AVEC 2014 datasets. Our code is publicly available at \href{https://github.com/PushineLee/CDP}{https://github.com/PushineLee/CDP}.

\end{abstract}


\section{Introduction}

\IEEEPARstart{D}{epression} is a prevalent mental disorder presented as persistent feelings of sadness, debilitation, and loss of interest in activities \cite{world2017depression, aprilia2022investigating, marwaha2023novel}. It increases the risk of suicide and accounts for a substantial psychological burden \cite{bachmann2018epidemiology, singh2023effectiveness}. The existing primary diagnostic methods used involve mental health reports, such as the Beck Depression Inventory (BDI-II) \cite{beck1996beck} (target variable focused on in this paper), the Hamilton Depression Rating Scale (HRSD) \cite{hamilton1986hamilton}, and the Patient Health Questionnaire (PHQ-8) \cite{kroenke2009phq}. The diagnostic process depends on interviews and demands considerable time and effort from both psychiatrists and patients. Also, it heavily relies on the clinicians' subjective experience, as well as the patients' cognitive abilities and psychological states.  

Over the past decade, continuous efforts have been made to explore effective biomarkers and methods for automated depression prediction. For depression biomarkers, initial attention was focused on hand-crafted features such as Local Binary Patterns (LBP) \cite{shan2009facial}, Local Gabor Binary Patterns from Three Orthogonal Planes (LGBP-TOP) \cite{almaev2013local}, and Local Binary Pattern from Three Orthogonal Planes (LBP-TOP) \cite{dhall2015temporally,wen2015automated} extracted from facial videos. Handcrafted features extracted from other behavioral signals have also been employed in this task, including speech, facial action units, facial landmarks, head poses, and gazes \cite{williamson2014vocal,du2019encoding}. The development of these hand-crafted features heavily relies on specific knowledge related to the depression prediction tasks. Acquiring such knowledge is time-consuming, subject to researchers' subjective cognition and task specificity, and lacks good generalization. Moreover, these features struggle to identify certain implicit and difficult-to-distinguish patterns of depression \cite{he2022deep}. Fortunately, with the advent of deep learning \cite{krizhevsky2012imagenet}, a new pathway has been opened for depression prediction tasks. In this pipeline end-to-end deep neural networks can be trained using depression data including facial video \cite{de2020deep,de2021mdn,zhu2017automated,9187982}, speech \cite{he2018automated,zhao2020hybrid,williamson2014vocal} and other data sources. These deep models are expected to discover subtle, indistinguishable depression-related features and implicitly discriminate among them for improved prediction.  

\begin{figure}[!t]
\centering
\includegraphics[width=0.5\textwidth]{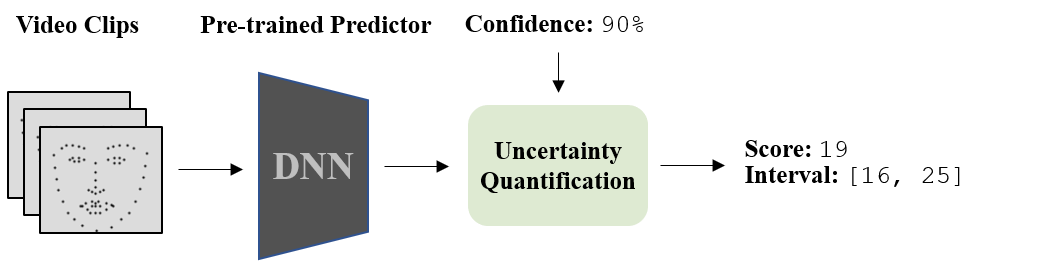}
\caption{Uncertainty quantification for facial depression prediction can provide an additional confidence interval to score prediction at any given confidence level, such as 90\%\protect\footnotemark[1].}
\label{fig1}
\end{figure}

\footnotetext[1]{To protect personal privacy, we show facial landmarks instead of the original facial images.}

While the rapid development of deep learning and computer vision has led to exciting progress in depression prediction, the reliability and trustworthiness of deep learning models-considered black boxes \cite{castelvecchi2016can, buhrmester2021analysis, xu2023confidence}-have not been well addressed. These models may exhibit overconfidence in predicting failures, potentially leading to catastrophic risks.
For automated depression prediction, the accuracy performance should not be the only criterion determining the usefulness of a predictive model. The lack of statistically rigorous uncertainty quantification is a key factor undermining the reliability of depression prediction. Consider the application context of depression prediction. We are more willing to accept reliable predictions that  provide insight into potential risks. Beyond the depression severity prediction, we need additional outcomes to represent their uncertainty. As shown in Fig. \ref{fig1}, our aim is to provide statistically rigorous confidence intervals in depression prediction. These intervals should cover the actual depression scores with any user-defined level of confidence.

In this work, we propose conformal depression prediction (CDP) to quantify the uncertainty of depression severity estimation based on conformal prediction (CP \cite{vovk1999machine,shafer2008tutorial}). CDP can provide confidence intervals to satisfy marginal coverage for the actual depression scores. We show that, CDP as a plug-and-play module, can validly quantify uncertainty for most depression prediction models, without retraining the model. Furthermore, inspired by the \textit{adaptive prediction sets} (APS) \cite{romano2020classification} in uncertainty-aware classification tasks, which construct a prediction set per-input adaptive to its prediction distribution, we propose conformal depression prediction with approximate conditional coverage (CDP-ACC) to overcome the shortcomings of conformal prediction with marginal coverage, providing adaptive confidence intervals for depression prediction.

The main contributions are summarized as follows:

1. We propose an uncertainty quantification framework CDP for facial depression prediction. CDP provides statistically rigorous confidence intervals that satisfy marginal coverage for most existing facial depression prediction models. 

2. We develop CDP-ACC, an improved CDP algorithm that satisfies approximate conditional coverage, yielding tighter confidence intervals while maintaining good coverage. 

3. We conduct extensive experiments to demonstrate the effectiveness of our proposed method in uncertainty quantification of facial depression prediction. 

\section{RELATED WORK}
\subsection{Depression Prediction} 
Over the past decade, depression prediction based on deep learning has attracted significant interest and attention in the field, leading to many meaningful research advancements. Studies using labeled depression data from various modalities, including speech, text, video, and EEG signals, as well as the integration of these modalities, have taken a leading role. 

For single-modality depression prediction, He et al. employed a Deep Convolutional Neural Network (DCNN) to extract depression-related features from speech data \cite{he2018automated}. These learned features were subsequently integrated with additional hand-crafted features to predict depression scores. Zhao et al. introduced a hybrid network that integrates Self-Attention Networks (SAN) and DCNN to handle both low-level acoustic features and 3D Log-Mel spectrogram data \cite{zhao2020hybrid}. Following this, feature fusion techniques along with support vector regression are employed for depression prediction. Melo et al. concentrated on efficiently extracting spatiotemporal features from facial videos associated with depression \cite{de2020deep,de2021mdn}. They introduced a deep Multiscale Spatiotemporal Network (MSN) and a deep Maximization-Differentiation Network (MDN) to predict depression levels. Sharma et al. proposed CNN-LSTM hybrid networks to extract depression features from EEG signals \cite{sharma2021dephnn}. Seal et al. developed the DeprNet, a DCNN designed for classification of EEG data from depressed and normal subjects \cite{9335239}. 

For multimodal depression prediction, Niu et al. introduced a novel Spatio-Temporal Attention (STA) network and a Multimodal Attention Feature Fusion (MAFF) strategy to acquire a multimodal representation of depression cues from facial and speech spectrum  \cite{niu2020multimodal}. Uddin et al. introduced a novel deep multi-modal framework that effectively integrates facial and verbal cues for automated depression prediction \cite{uddin2022deep}. They extracted features separately from video and speech and employed a Multi-modal Factorized Bilinear pooling (MFB) strategy to efficiently fuse these multi-modal features. Niu et al. proposed a novel multi-layer perceptron (MLP) architecture named DepressionMLP, designed to automatically predict depression levels through facial keypoints and action units (FKRS and AURS) \cite{niu2024depressionmlp}.

These studies, either using single-modality data or multimodal data, primarily focus on the prediction accuracy, while overlooking the analysis of predictive uncertainty. Interestingly, works such as those of Melo et al. introduced minimizing expected error to learn distribution of depression levels \cite{de2019depression}, while Zhou et al. proposed DJ-LDML to learn depression representation using a label distribution and metric learning \cite{9187982}. These studies implicitly explore uncertainty in depression prediction, yet without providing a clear and intuitive uncertainty quantification. The method closely related to ours is the one proposed by Nouretdinov et al. \cite{nouretdinov2011machine}, which leveraged conformal prediction for uncertainty quantification of MRI-based depression classification. Different from the work \cite{nouretdinov2011machine} that dedicated to the classification task, we focus on regression task in the context of deep learning, and we further develop an approximate conditional coverage (ACC) algorithm to overcome the shortcomings of conformal prediction with marginal coverage.

\subsection{Uncertainty Estimation in Affective Computing}
Uncertainty has been widely considered for various tasks in affective computing, such as emotion recognition, facial expression recognition, and apparent personality recognition \cite{harper2020bayesian,prabhu2023end,wu2023estimating,lo2023modeling,she2021dive,le2023uncertainty,tellamekala2022dimensional}. In emotion recognition, Harper et al. utilized a Bayesian framework to support emotional valence classification from the perspective of uncertainty \cite{harper2020bayesian}. Prabhu et al. proposed BNNs to model the label uncertainty based on subjectivity in emotion recognition \cite{prabhu2023end}. Wu et al. employed deep evidence regression to jointly model aleatoric and epistemic uncertainties, aiming to enhance the performance of emotion attribute estimation \cite{wu2023estimating}. Tellamekala et al. introduced calibrated and ordinal latent distribution to capture modality-wise uncertainty for multimodal fusion in emotion recognition \cite{tellamekala2023cold}. Lo et al. utilized probabilistic uncertainty learning to extract information from low-resolution data and proposed the emotion wheel to learn label uncertainty, yielding robust facial expression recognition \cite{lo2023modeling}. In facial expression recognition, She et al. adopted a pairwise uncertainty estimation approach, assigning lower confidence scores to more ambiguous samples to address annotation ambiguities \cite{she2021dive}. Le et al. proposed a method based on uncertainty-aware label distribution to adaptively construct the training sample distribution \cite{le2023uncertainty}. Additionally, Tellamekala et al. utilized a neural latent variable model to model aleatoric and epistemic uncertainty and integrated both to enhance the performance of apparent personality recognition \cite{tellamekala2022dimensional}.  

The uncertainty-inspired algorithms mentioned above may not be suitable for depression prediction, as the available datasets with annotated depression scores are limited. Re-stabilizing a well-performing, uncertainty-inspired depression prediction algorithm is not straightforward. Furthermore, while the above methods may achieve better accuracy performance in their respective tasks by introducing uncertainty-inspired heuristics, they do not establish a rigorously valid uncertainty quantification. In contrast to previous methods, our proposed CDP is a plug-and-play uncertainty quantification method for depression prediction, which does not require model retraining and does not assume the depression data distribution. Thanks to the conformal prediction \cite{vovk1999machine,vovk2012conditional}, our CDP can achieve valid confidence intervals with theoretical coverage guarantees for depression predictions.  

\begin{figure}
\centering
\includegraphics[width=0.5\textwidth]{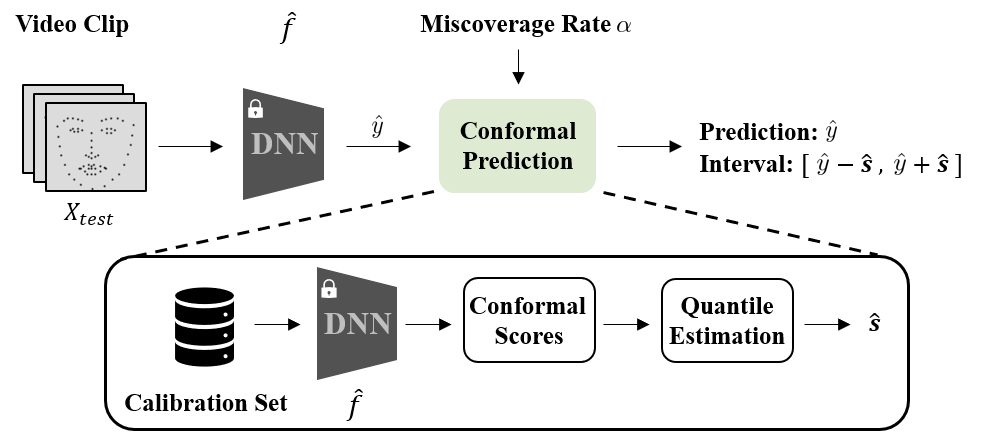}
\caption{The pipeline of CDP. CDP takes as input a pre-trained depression prediction model $\hat{f}$, a calibration set $\mathcal{D}_{cal}$ and a given confidence level $1 - \alpha$. CDP sets the prediction biases (on the calibration set) as the conformal scores and employs the $(1 - \alpha)$-th empirical quantile of the conformal scores to form confidence interval for the prediction $\hat{f}(X_{test})$.}
\label{fig2}
\end{figure}

On the other hand, current conformal prediction methods tend to produce unnecessarily conservative intervals, resulting in the confidence interval to be unnecessarily wide \cite{romano2019conformalized}. In depression prediction, overly wide intervals are meaningless. For instance, the prediction with BDI-II (ranging from 0 to 63) has a 100\% confidence interval of [0, 63]. Clearly, such a predictive uncertainty lacks practicality. 
Additionally, marginal coverage guarantees only valid average intervals, which overcovers simple subgroups and undercovers difficult ones \cite{angelopoulos2021gentle}. Differently, our proposed conformal prediction method with approximate conditional coverage (CDP-ACC) can produce valid and tighter confidence interval adaptive to the specific depression prediction.

\section{METHOD}
\subsection{Preliminaries}
Formally, let us assume that we have the facial depression-associated video data denoted by $X$ as input and the corresponding label denoted by $y$ (the BDI-II score \cite{beck1996beck}, for example). We randomly split the data into disjoint training set $\mathcal{D}_{train}$, calibration set $\mathcal{D}_{cal} = \{(X_{i},y_i)\}_{i=1}^{N}$, and test set $\mathcal{D}_{test}$. We use $\mathcal{D}_{train}$ to train a DNN model $\hat{f}(\cdot)$ for depression prediction. Here, $\hat{y} \in \mathbb{R}$ denotes the point prediction for $X$ by $\hat{f}(\cdot)$. In the deep learning context, we can formulate the depression prediction as a deep regression task by minimizing the mean squared error (MSE) loss as the learning objective. 

\begin{equation}\label{mse}
  \begin{split}
  \mathcal{L}_{MSE} = \frac{1}{2|\mathcal{D}_{train}|}\sum_{i=1}^{|\mathcal{D}_{train}|}(\hat{y}_i - y_i)^2
  \end{split}
  \end{equation}
where $|\mathcal{D}_{train}|$ denotes the number of data in $\mathcal{D}_{train}$.

In depression prediction based on such regression models, the point prediction $\hat{y}$ is the expectation $\mathbb{E}[y|X]$ derived from estimating the conditional probability distribution $P(y|X)$. However, we do not have access to the oracle of $P(y|X)$. 
To quantify the uncertainty of depression prediction $\hat{y}$, we expect to construct an interval function $\mathcal{C}_{\alpha}(X)$, which takes depression data $X$ and miscoverage rate $\alpha$ as input and outputs a confidence interval. The interval contains the depression target $y$ with any confidence level $1 - \alpha$, as shown in Fig. \ref{fig1} ($\alpha = 0.1$). Clearly, wider interval $|\mathcal{C}_{\alpha}(X)|$ indicates greater uncertainty for the model's prediction $\hat{y}$.

\subsection{Depression Prediction with UQ}

The straightforward method for uncertainty quantification of depression prediction is to train a DNN to estimate the conditional probability distribution $P(y|X=x)$ of depressed individuals. Given our lack of knowledge regarding the true $P(y|X=x)$, one possible solution is to assume that the depression target $y$ follow a Gaussian distribution $y \sim \mathcal{N}(\hat{y},\sigma^2 )$, and uses the negative log-likelihood (NLL) \cite{cui2020calibrated} as the loss function:
\begin{equation}\label{NLL}
    \mathcal{L}_{NLL} = \sum_{i = 1}^{|\mathcal{D}_{train}|}\frac{\text{log}\sigma_i^2}{2} + \frac{(y_i - \hat{y}_i)^2}{2\sigma_i^2}
\end{equation}
The learning objective allows us to train a DNN capable of predicting the distribution $P(y|X=x)$ for any given individual $x$. Furthermore, under the Gaussian assumption, the $\alpha$-th quantile can be calculated by
\begin{equation}\label{gaussian_q}
    \hat{y}_{\alpha} = \hat{y} + \sqrt{2}\sigma \cdot \mathrm{erf}^{-1}(2\alpha-1) 
\end{equation}
where $\mathrm{erf}^{-1}$ denotes the inverse Gaussian error function. In this way, we can obtain the confidence interval $[\hat{y}_{\alpha/2}, \hat{y}_{1 - \alpha/2}]$ that satisfies $P(y \in [\hat{y}_{\alpha/2}, \hat{y}_{1 - \alpha/2}]) \geq 1 - \alpha$. 

In practice, however, depression datasets (e.g., AVEC 2013 and AVEC 2014 \cite{valstar2013avec,valstar2014avec}) exhibit high data imbalance, and the assumption of the Gaussian distribution may not be appropriate in this context. We prefer a distribution-free method for uncertainty quantification. Indeed, by introducing the pinball loss (Eq. (\ref{pinball})), we can train a DNN as a conditional quantile estimator \cite{taylor2000quantile,tagasovska2019single}, allowing us to obtain arbitrary quantile values for $P(y|X=x)$:

\begin{equation}\label{pinball}
\mathcal{L}_{pinball} =
\begin{cases}
q \cdot (y - \hat{y}_q), & \text{if } y \geq \hat{y}_q \\
(1 - q) \cdot (\hat{y}_q - y), & \text{if } y < \hat{y}_q
\end{cases}
\end{equation}
where $ q \in [0, 1]$, $\hat{y}_q$ denotes the $q$-th quantile of the prediction, and $y$ is the ground truth. Using the trained regressor, we can use ${\alpha/2}$ and ${1 - \alpha/2}$ quantile values of the predictive distribution to derive the confidence interval $[\hat{y}_{\alpha/2}, \hat{y}_{1 - \alpha/2}]$, thereby achieving a distribution-free uncertainty quantification. 

\begin{algorithm}[t]
\caption{Conformal Depression Prediction} 
\label{cp}
\SetAlgoLined
\KwIn{$\mathcal{D}_{cal} = \{X_{i},y_i\}_{i = 1}^{N}$: calibration set; \\ 
$\hat{f}(\cdot)$: pre-trained depression prediction model; \\
$X_{test}$: test data; \\
$\alpha \in (0,1)$: miscoverage rate; \\ 
$\mathcal{Q}(\cdot,\cdot)$: quantile estimator }
\KwOut{Confidence interval $\mathcal{C}_{\alpha}(X_{test})$}

\tcp{Calculate the conformal scores $\{s_i\}_{i = 1}^{N}$ on $\mathcal{D}_{cal}$} 
\For{$i = 1$ \KwTo $N$}{$\hat{y}_i = \hat{f}(X_i)$ \\ $s_i = |\hat{y}_i - y_i |$}
    
$q = \lceil \frac{N+1}{N} \rceil (1 - \alpha)$

\tcp{Calculate the $q$-th quantile of the conformal scores}

$\hat{s} = \mathcal{Q}(\{s_i\}_{i = 1}^{N},q)$ 

\tcp{Calculate the confidence interval}
$\mathcal{C}_{\alpha}(X_{test}) = [\hat{f}(X_{test}) - \hat{s}, \hat{f}(X_{test}) + \hat{s}]$ 

\Return $\mathcal{C}_{\alpha}(X_{test})$
\end{algorithm}

However, due to limited depression data and imbalanced data distributions, using Eq. (\ref{NLL}) or Eq. (\ref{pinball}) to train depression prediction models is challenging. In practice, it has been observed that those methods suffer from overconfidence, where $P(y \in \mathcal{C}_{\alpha}(X))$ can be significantly lower than $1 - \alpha$. In this context, there is an urgent need for a post-processing method that is distribution-free to quantify the uncertainty of depression prediction. In particular, we seek for constructing the confidence interval $\mathcal{C}_{\alpha}(\cdot)$ to guarantee the marginal coverage so as to maintain an average coverage rate across the target, as expressed by

\begin{equation}\label{marginal_demo}
    P(y_{test} \in \mathcal{C}_{\alpha}(X_{test})) \geq 1 - \alpha, \  \ \   (X_{test}, y_{test}) \in \mathcal{D}_{test}
\end{equation}

To that end, we introduce conformal prediction (CP \cite{vovk1999machine,sesia2021conformal}) for uncertainty quantification of facial depression prediction. Specifically, we develop conformal depression prediction (CDP) as shown in Fig. \ref{fig2}. As premise of CP is to have a calibration set that is exchangeable with the test sample (This is guaranteed if the i.i.d. assumption holds), we calculate the conformal score $s$ on the calibration set, 

\begin{equation}\label{CDP_score_0}
   s_i = |y_i - \hat{y}_i|
\end{equation}
which reflects the discrepancy between the depression prediction and the ground truth. Next, we calculate the $q$-th quantile of $s$ and denote it as $\hat{s}$. For any test sample $X_{test}$, we have $P(s_{test}\le \hat{s}) = q$. In this way, we can obtain its prediction intervals $\mathcal{C}_{\alpha}(X_{test}) = [\hat{y}_{test} - \hat{s}, \hat{y}_{test} + \hat{s}]$, which is valid with 

\begin{equation}\label{CDP_score}
   1 - \alpha \le P(y_{test} \in \mathcal{C}_{\alpha}(X_{test}) \le 1 - \alpha + \frac{1}{N+1}
\end{equation}

The implementation of CDP are summarized in Algorithm \ref{cp}. With CDP, we can acquire valid, theoretical coverage guaranteed confidence interval, and the interval width relates to the user-defined confidence levels $1 - \alpha$. Note that this CP-based uncertainty quantification achieves marginal coverage for the target depression scores at any confidence levels, as proven in \cite{vovk1999machine,sesia2021conformal}.


\subsection{CDP with Approximate Conditional Coverage}

There is a stronger notion than the marginal coverage, namely conditional coverage, as defined by 
\begin{equation}\label{conditional}
    P(y_{test} \in \mathcal{C}_\alpha(y|X_{test}))\geq 1 - \alpha
\end{equation}
For the model prediction $\hat{y}_{test}$ on any test sample $X_{test}$, it aims to provide a confidence interval $C_{\alpha}$ with $1 - \alpha$ coverage. 
In contrast to marginal coverage, conditional coverage is more suitable for uncertainty quantification in individual depression prediction. In practical applications, each individual only cares about whether the prediction is trustworthy (or within which confidence interval the prediction falls). 
Unfortunately, achieving conditional coverage is practically challenging in a distribution-free setting \cite{vovk2012conditional}. Our work draws inspiration from ordinal adaptive prediction sets \cite{lu2022improving} for classification tasks. The motivation behind our CDP-ACC is to generate adaptive confidence intervals for pre-trained depression predictor, achieving approximate conditional coverage in a model-agnostic and distribution-free manner. \newline


\begin{figure}
\centering
\includegraphics[width=0.55\textwidth]{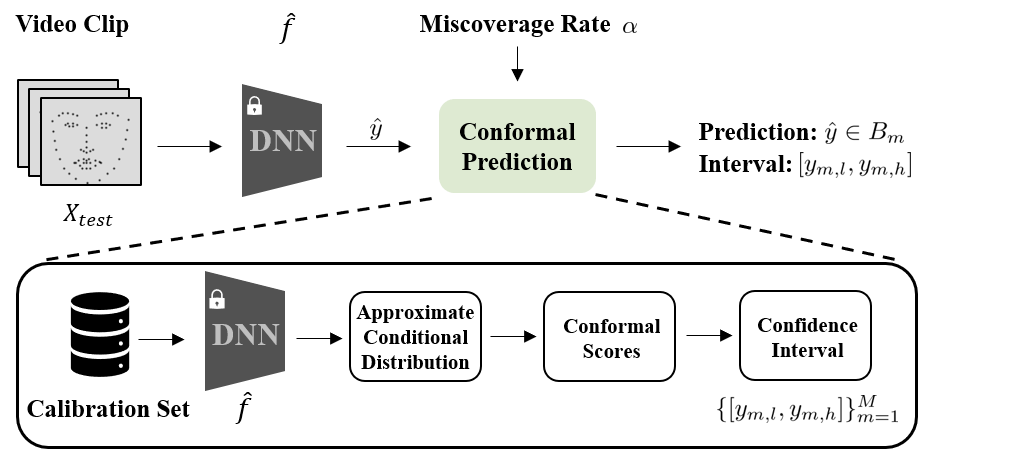}
\caption{The pipeline of CDP-ACC. The predictions on the calibration set are firstly partitioned into $M$ subintervals. For each subinterval, CDP-ACC estimates the approximate conditional distribution $P(y|\hat{y})$, computes the conformal score based on the calibration samples in that interval, and finally constructs the adaptive confidence interval. }
\label{fig_CDP-ACC}
\end{figure}

\subsubsection{Approximate Conditional Coverage} 

Given a pre-trained depression predictor with point prediction, $\hat{y} = \hat {f}(X)$, we cannot directly access to the conditional distribution $P(y|X)$. 
To accurately estimate $P(y|X)$ in depression prediction, we rely on a key assumption  \cite{chung2021beyond}: similar $X$ (in feature embedding space) will have similar conditional distributions $P(y|X)$ and similar conditional expectation. This assumption is reasonable in our practical observations. For example, different video clips of the same patient tend to have similar depression predictions. 

Let $\Delta(X_i)$ denote the neighborhood of $X_i$ in the feature embedding space. Our aim is to approximate $P(y|X=X_i)$ with $P(y|X \in \Delta(X_i))$. In the context of facial depression prediction, $X_i$ denotes the facial video clips, and $\Delta(X_i)$ can represent clips belonging to the same video, or clips with the same depression score. However, for unseen test video clips, we do not have oracle with access to $\Delta(X_i)$. We need to further explore relaxation scheme to approximate $\Delta(X_i)$.

We revisit $\Delta(X_i)$ from the perspective of the model predictions (with pre-trained model $\hat{f}$): $\Delta(\hat{y_i}) = \hat{f}(\Delta(X_i))$. The model predicts similar depression scores for what it perceives as similar $X_i$. Thus, we can use $\Delta(\hat{y_i})$ as the condition instead of $\Delta(X_i)$, that is, similar predictions $\hat{y}$ can also have similar conditional distributions. With this approximation, we have $P(y|X = X_i) \approx P(y| \hat{y}\in \Delta(\hat{y}))$, and we can estimate the conditional distribution by grouping neighboring points of the depression predictions. Moreover, by incorporating the approximation, we can introduce \textit{inductive bias}, which implies that for samples falling in $\Delta(\hat{y})$, the more concentrated the points, the higher the credibility.

\begin{figure}
\centering
\includegraphics[width=0.5 \textwidth]{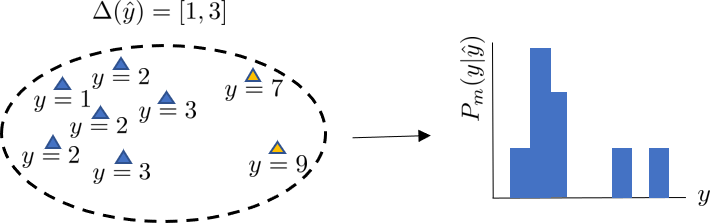}
\caption{Histogram estimation of the conditional distribution $P(y| \hat{y}\in \Delta(\hat{y}))$. For samples falling in $\hat{y} \in \Delta(\hat{y})$, the more concentrated the points, the higher the credibility. }
\label{fig3}
\end{figure}

We uniformly divide the range of $\hat{y}$ into $M$ subintervals $B_m,\ m = 1,..., M$. Let $y_{m,L}$ and $y_{m,U}$ denote the lower and upper bounds of $y$ in each subinterval $m$. Based on the approximation $P(y|X = X_i) \approx P(y|\hat{y} \in B_m)$, we can only use the depression label $y$ (corresponding to $\hat{y}$ within $B_m$) to approximately estimate the conditional distribution $P(y|X = X_i)$. Following \cite{sesia2021conformal,lu2022improving}, we use histogram binning to estimate the conditional probability distribution $P_{m}(y|\hat{y} \in B_m)$ for each subinterval $m$ (shown in Fig. \ref{fig3}).\newline


\subsubsection{Confidence Interval} In conformal prediction, setting the conformal score is particularly important. The choice conformal score function essentially determines the effectiveness of the conformal prediction method. In CDP-ACC, setting the conformal score involves constructing nested sequence of approximate oracle intervals. 

\begin{figure}
\centering
\includegraphics[width=0.28 \textwidth]{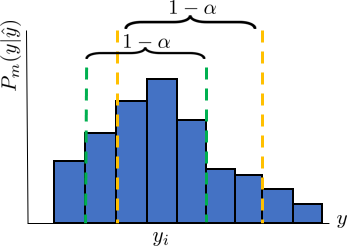}
\caption{The setting of the conformal score for CDP-ACC. Given the approximate conditional distribution $P(y|\hat{y})$, various prediction intervals can be acquired containing $y_i$ with $1 - \alpha$ coverage, of which the width of the shortest one is set as the conformal score for our CDP-ACC.}
\label{Dep_CP_movation}
\end{figure}

To maintain the continuity of prediction interval, we construct a nested sequence $E(\cdot)$ so that the interval with higher confidence covers the interval with lower confidence, all while endeavoring to minimize the width of each interval. As shown in Fig. \ref{Dep_CP_movation}, for each $y_i$ whose prediction falls into $B_m$ we can construct the confidence interval $E_m(\tau, y_i)$ with the shortest width:

\begin{equation}\label{score}
\begin{aligned}
E_m(\tau, y_i)=\mathop{\arg\min}\limits_{\substack{[y_{l},y_{h}] \subset [y_{m,L},y_{m,U}], \\ y_i \in [y_{l},y_{h}]}} & \{y_{h}  - y_{l} : \\ 
& F_{m}(y_h|\hat{y}) - F_{m}(y_l|\hat{y}) \geq \tau\}
\end{aligned}
\end{equation}
where $\tau \in (0,1)$ denotes the confidence level, $m \in [1,M]$ and $F_{m}$ denotes the cumulative probability distribution of the conditional probability distribution $P_{m}(y|\hat{y} \in B_m)$. If the optimal solution is not unique, we can add a little noise $\epsilon$ into the $B_m$ to break the ties \cite{barber2021predictive}. It should be noted that, given $y_i$, $E_{m}(\tau, y_i)$ is a nested sequence as $\tau$ grows, for example, $E_m(0.1,y_i) \subseteq E_m(0.2,y_i) \subseteq E_m(0.3,y_i)$. For any sample $y_i$ whose prediction falls into $B_m$, we set the width of its interval $E_m$ as the conformal score $s_i$:  

\begin{equation}\label{new_score}
s_i(m,\alpha) = |E_m(1 - \alpha,y_i)|
\end{equation}
The conformal score typically reflects the prediction bias. Higher scores indicate poorer prediction quality while wider interval corresponds to worse predictions in our context.

After calculating the conformal scores on the calibration set, we can estimate the confidence interval for any test samples based on the assumption that the calibration set and the test samples are exchangeable to generate valid confidence intervals. A good confidence interval should achieve the specified coverage rate at least. Narrower interval widths are desirable as they provide more precise uncertainty quantification. For $E_m(\tau,y_i)$, if the user sets the confidence level $\tau$ to 100\%, the confidence interval for any samples $y_i$ will be unique ($E_m(1,y_i)$). Otherwise, there will be multiple prediction intervals that cover $y_i$ with at least $1 - \alpha$ probability. According to our smoothness approximation, the concentration degree of samples within a subinterval can represent prediction uncertainty. We use interval width as the conformal score and identify the narrowest interval that covers $y_i$ with $1 - \alpha$ probability as its confidence prediction interval.
Specifically, given all conformal scores calculated on the calibration set and a user-defined miscoverage rate $\alpha$, CDP-ACC constructs confidence interval for any unseen test sample $X_{test}$ according to the following steps. First, for each subinterval $m$, we calculate the $(1 - \alpha)$-th quantile of the conformal scores $\{s_i(m,\alpha)\}$. 

\begin{equation}
\label{qe}
 \hat{s}(m,\alpha) = \mathcal{Q}(\{s_i(m,\alpha)\},1 - \alpha)
\end{equation}

Then, for each subinterval $m$ we can obtain its confidence interval with shortest width according to 
\begin{equation}
\begin{aligned}\label{interval}
\mathcal{A}(m,\alpha)=\mathop{\arg\min}\limits_{\substack{[y_{l},y_{h}] \subset [y_{m,L},y_{m,U}], \\ y_i \in [y_{l},y_{h}]}}\{y_{h} - y_{l} : 
s_i(m,\alpha) \leq \hat{s}(m,\alpha) \}
\end{aligned}
\end{equation}

Finally, based on the interval into which the prediction of the test sample falls, we can retrieve the corresponding confidence interval. The proposed CDP-ACC is summarized in Algorithm \ref{CP-ACC}.

\begin{algorithm}[t]
\caption{Depression Prediction with CDP-ACC} 
\label{CP-ACC}
\SetAlgoLined
\KwIn{$\mathcal{D}_{cal} = \{X_{i},y_i\}_{i = 1}^{N}$: calibration set; \\
$\hat{f}(\cdot)$: pre-trained depression prediction model; \\
$X_{test}$: test data; \\
$\alpha \in (0,1)$: miscoverage rate; \\
$L,U$: lower and upper bounds of the predictions on $\mathcal{D}_{cal}$; \\
$\mathcal{H}(\cdot)$: histogram estimator; \\
$\mathcal{Q}(\cdot,\cdot)$: quantile estimator; \\
$M$: number of the subintervals}  

\KwOut{Confidence interval $\mathcal{C}_{\alpha}(X_{test})$}
Initialize subintervals $\{B_m\}_{m = 1}^{M} \gets \emptyset$

\tcp{Subintervals dividing for $\hat{y}$ on $\mathcal{D}_{cal}$}

\For{$i = 1$ \KwTo $N$}{ 
$\hat{y}_i = \hat{f}(X_i)$ 

\For{$m = 1$ \KwTo $M$}{
 \If{$\hat{y}_i \in [L + \frac{m-1}{M}(U-L), L + \frac{m}{M}(U-L))$}{
 $B_m \cup \{y_i\}$}}}
 
 \tcp{Estimation of the conditional distribution}
\For{$m = 1$ \KwTo $M$}{
 $P_{m}(y|\hat{y})$ = $\mathcal{H}(B_m)$ \\
 $F_{m}(y|\hat{y})$ = $\int_{y_{m,L}}^yP_{m}(t|\hat{y})dt$
 }

 \tcp{Calculate the confidence interval}
  
 $\hat{y}_{test} = \hat{f}(X_{test})$ \\
\For{$m = 1$ \KwTo $M$}{
 $N_m = |B_m|$\\
 \tcp{Calculate conformal score according to Eq. (\ref{score}) and Eq. (\ref{new_score})}
 $\{s_i(m,\alpha)\}_{i = 1}^{N_m} = \{|E_m(1 - \alpha,y_i)|\}_{i = 1}^{N_m}$ \\
 
\tcp{Calculate the ($1 - \alpha$)-th quantile of the conformal scores}
 $\hat{s}(m,\alpha) = \mathcal{Q}(\{s_i(m,\alpha)\}_{i=1}^{N_m},1 - \alpha)$ \\
 \tcp{Calculate the interval according to Eq. (\ref{interval})}
 $[y_{m,l}, y_{m,h}] = \mathcal{A}(m,\alpha)$\\
 } 
\If{$\hat{y}_{test} \in B_m$}{
 $\mathcal{C}_{\alpha}(X_{test}) = [y_{m,l}, y_{m,h}]$ \\
 \Return $\mathcal{C}_{\alpha}(X_{test})$
 }
\end{algorithm}

\section{EXPERIMENTS}

\subsection{Dataset}
In order to verify the effectiveness of our uncertainty quantification methods in depression prediction, we conduct experiments on two commonly used facial depression datasets, AVEC 2013 \cite{valstar2013avec} and AVEC 2014 \cite{valstar2014avec}. 
AVEC 2013 is a subset of the AViD-Corpus, consisting of 150 facial videos. The videos range in length from 20 to 50 minutes, with a frame rate of 30 FPS and a resolution of 640 $\times$ 480 pixels. The content of the videos includes facial recordings of subjects during human-computer interaction tasks, such as reading assigned content and improvising on given sentences. We divide the dataset into 70 videos for training, 30 videos for validation (calibration), and 50 videos for testing. The facial depression levels are annotated using BDI-II scores \cite{beck1996beck}, with each video assigned a single value representing the depression level. The BDI-II questionnaire consists of 21 questions, each with multiple-choice answers scored from 0 to 3. The total BDI-II score ranges from 0 to 63: 0-13 indicates minimal depression, 14-19 indicates mild depression, 20-28 indicates moderate depression, and 29-63 indicates severe depression, as shown in Table \ref{bdi}.

AVEC 2014, a subset of AVEC 2013, consists of two tasks: Northwind and Freeform. It comprises 150 Northwind videos and 150 Freeform videos. The original videos of AVEC 2014 are similar to those used in the AVEC 2013 but include five pairs of previously unseen videos, replacing a few videos considered unsuitable. We use 200 videos for training, 50 videos for validation (calibration), and 50 videos for testing.

The videos in the AVEC 2014 dataset are relatively short, with an overlap of 8 frames between adjacent video clips, whereas there is no overlap in the AVEC 2013 dataset. During data processing, landmarks in all video frames are detected using OpenFace \cite{baltruvsaitis2016openface}, and facial region alignment is performed using the eyes, nose, and mouth.

\subsection{Experimental Setup}
\subsubsection{Prediction Model}
This work focuses on methods for quantifying uncertainty in depression prediction rather than on the predictive performance of the models themselves. We use the classic C3D \cite{tran2015learning,zhu2017automated} and the SlowFast \cite{feichtenhofer2019slowfast} networks as baseline models for facial depression prediction. In particular, for vanilla regression, we employ MSE loss as the learning objective during training and modify the last FC layer from 4096 to 64 to enhance training stability and prevent overfitting. In the context of Quantile Regression (QR \cite{taylor2000quantile}), we use 99 quantiles evenly spaced between 0.01 and 0.99 for the pinball loss. The output of the last FC layer transforms from a single output in vanilla regression to 99 outputs in QR.\newline 

\subsubsection{UQ for Depression Prediction} 
To quantify the uncertainty of depression prediction models, we perform NLL \cite{cui2020calibrated} and QR \cite{taylor2000quantile} that achieves UQ by retraining the model. 
We also implement several post-processing methods related to CP \cite{vovk1999machine,vovk2012conditional} to construct valid confidence intervals for depression prediction. 
Specifically, CQR \cite{romano2019conformalized} is a post-processing method that applies conformal prediction to calibrate the interval of pre-trained QR. Its conformal score is set as $max\{\hat{f}(X; \alpha /2) - y, y - \hat{f}(X; 1 - \alpha /2)\}$. CDP represents the method for uncertainty quantification in vanilla regression, as shown in Fig. \ref{fig2}. Its conformal score is set as $|y - \hat{y}|$, which achieves marginal coverage. CDP-ACC is a method that achieves approximate conditional coverage and provides valid and adaptive confidence intervals.
We set the miscoverage rate $\alpha$ to 0.1. Given that the depression score ranges from 0 to 63, we use truncation operator to ensure the predictive interval $[\hat{y}_l, \hat{y}_h] \subseteq [0, 63]$. For CDP-ACC, we set the subintervals number $M$ to 14, lower bound $L$ to 0 and upper bound $U$ to 63. To validate the generalizability of our method, we also evaluate our UQ method on seven publicly available datasets for general regression task (for more information, please refer to the \textbf{Appendix}).\newline

\subsubsection{Metrics}
In depression prediction, mean absolute error (MAE) and root mean squared error (RMSE) are commonly used as the evaluation metrics, aiming to quantify the accuracy performance. To further measure the inherent uncertainty in predictions, we adopt the prediction interval coverage probability (PICP) and mean prediction interval width (MPIW) \cite{tagasovska2019single}. PICP and MPIW are two seemingly contradictory metrics. 
For example, if the PICP worsens, the MPIW may actually improve.
We aim to minimize MPIW while ensuring PICP coverage rate. If we cannot ensure that PICP is not less than ($1 - \alpha$) $\times$ 100\%, merely shortening the interval width is meaningless. Similarly, constructing an interval with a larger width to ensure coverage is also meaningless. \newline
\begin{equation}\label{picp}
PICP = \frac{1}{|\mathcal{D}_{test}|}\sum_{i = 1}^{|\mathcal{D}_{test}|}\mathbf{1}(y_i\in\mathcal{C}_{\alpha}(X_i))
\end{equation}
\begin{equation}\label{mpiw}
MPIW =  \frac{1}{|\mathcal{D}_{test}|}\sum_{i = 1}^{|\mathcal{D}_{test}|}|\hat{y}_{i,h} - \hat{y}_{i,l}|
\end{equation}
where $X_i \in \mathcal{D}_{test}$ and $[\hat{y}_{i,l}, \hat{y}_{i,h}] = \mathcal{C}_{\alpha}(X_i)$.

\begin{table}[t]
\centering
\caption{Prediction errors of the regression models with different training losses and backbones}
\label{baseline}{
\begin{tabular}{c c c c c}   
\hline
 Method & Backbone & Dataset & MAE & RMSE \\
\hline
MSE & C3D & AVEC 2013 &  7.14 & 8.97 \\ 
NLL \cite{cui2020calibrated}& C3D & AVEC 2013 &  7.02 & 8.92 \\ 
QR \cite{taylor2000quantile}& C3D & AVEC 2013 & \textbf{6.82} & \textbf{8.64} \\ \hline
MSE & C3D & AVEC 2014 & \textbf{6.54} & 8.32 \\
NLL \cite{cui2020calibrated} & C3D & AVEC 2014 &  7.10 & 9.24 \\ 
QR \cite{taylor2000quantile}& C3D & AVEC 2014 &  6.65 & \textbf{8.24} \\ \hline
MSE & SlowFast & AVEC 2013 &  7.49 & 9.37 \\ 
NLL \cite{cui2020calibrated} & SlowFast & AVEC 2013 &  7.30 & 9.21 \\ 
QR \cite{taylor2000quantile}& SlowFast & AVEC 2013 &  \textbf{7.22} & \textbf{9.04} \\ \hline
MSE & SlowFast & AVEC 2014 &  7.00 & 8.70 \\
NLL \cite{cui2020calibrated} & SlowFast & AVEC 2014 &  \textbf{6.29} & \textbf{8.11} \\ 
QR \cite{taylor2000quantile}& SlowFast & AVEC 2014 &  6.89 & 8.65 \\ \hline
\end{tabular}}
\end{table}
\begin{table}
\centering
\caption{Uncertainty quantification performance of different methods}
\label{UQ}{
\begin{tabular}{c c c c c}   
\hline
Method & Backbone & Dataset &  PICP & MPIW \\
\hline
NLL \cite{cui2020calibrated} & C3D & AVEC 2013 & 74.33\% & 13.17 \\
QR \cite{taylor2000quantile}& C3D & AVEC 2013 &  17.61\% & 4.73 \\ 
CQR \cite{romano2019conformalized} & C3D & AVEC 2013 &  87.94\% & 23.32 \\ 
CDP  & C3D & AVEC 2013 &  91.78\% & 22.82 \\
CDP-ACC & C3D & AVEC 2013 &  \textbf{92.00\%} & \textbf{21.48} \\ \hline
NLL \cite{cui2020calibrated} & C3D & AVEC 2014 & 76.52\% & 14.15 \\
QR \cite{taylor2000quantile}& C3D & AVEC 2014 &  21.84\% & 5.35 \\ 
CQR \cite{romano2019conformalized} & C3D & AVEC 2014 &  87.14\% & 23.53 \\ 
CDP & C3D & AVEC 2014 &  91.53\% & 27.47 \\
CDP-ACC & C3D & AVEC 2014 &  \textbf{93.72\%} & \textbf{20.17} \\ \hline
NLL \cite{cui2020calibrated} & SlowFast & AVEC 2013 & 66.94\% & 17.69 \\
QR \cite{taylor2000quantile}& SlowFast & AVEC 2013 &  22.30\% & 5.38 \\ 
CQR \cite{romano2019conformalized} &SlowFast & AVEC 2013 &  92.29\% & 27.78 \\ 
CDP & SlowFast & AVEC 2013 & \textbf{95.25\%} & 28.29  \\ 
CDP-ACC & SlowFast & AVEC 2013 &  91.25\% &  \textbf{24.13}\\ \hline
NLL \cite{cui2020calibrated} & SlowFast & AVEC 2014 & 69.57\% & 19.59 \\
QR \cite{taylor2000quantile}& SlowFast & AVEC 2014 &  70.49\% & 16.83 \\ 
CQR \cite{romano2019conformalized} & SlowFast & AVEC 2014 &  93.06\% & 27.76 \\
CDP & SlowFast & AVEC 2014 &  \textbf{93.86\%} & 26.28 \\
CDP-ACC & SlowFast & AVEC 2014 &  92.18\% & \textbf{23.29} \\ \hline
$\alpha = 0.1$
\end{tabular}}
\end{table}

Furthermore, we utilize size-stratified coverage (SSC \cite{angelopoulosuncertainty, angelopoulos2021gentle}) to assess the extent to which different methods achieve conditional coverage for subsets of varying depression severity. We divide all possible test data into $G$ bins based on the severity of depression (shown in Table \ref{bdi}), denoted as $\mathcal{I}_1,  \mathcal{I}_2, \ldots, \mathcal{I}_G$. 
\begin{equation}\label{SSC_eq}
SSC = \underset{g \in \{1,..,G\}}{\min}\frac{1}{|\mathcal{I}_g|}\sum_{i \in \mathcal{I}_g}{\mathbf{1}\{y_i \in \mathcal{C}_{\alpha}(X_i)\}}
\end{equation}
where $\mathcal{I}_g \subset \{1, 2, \ldots, n_{\text{test}}\}$ represents the set of sample indices falling into bin $g$, $|\mathcal{I}_g|$ denotes the size of $\mathcal{I}_g$ and $\mathbf{1}(\cdot)$ denotes the indicator function.

\subsection{Results}

In this section, we provide a comprehensive evaluation of uncertainty quantification methods for depression prediction. Firstly, we show the prediction errors for vanilla regression, NLL and QR, allowing for a comparison of their accuracy performance in depression prediction. Then, we compare the uncertainty quantification performance of different UQ methods in depression prediction. We also perform uncertainty analysis of CDP-ACC for some instances to gain insight into CDP-ACC. Additionally, in the \textbf{Appendix}, we experimentally evaluate the proposed method on several general regression datasets.\newline

\begin{table}
\centering
\caption{Conditional coverage performance of different methods on AVEC 2014}
\label{SSC}{
\begin{tabular}{c c c c c}   
\hline
Method & Backbone & SSC \\
\hline
NLL \cite{cui2020calibrated} & C3D & 0.2426\\  
QR \cite{taylor2000quantile}& C3D & 0.1715 \\ 
CQR \cite{romano2019conformalized} & C3D & 0.8280 \\ 
CDP \cite{vovk1999machine}& C3D & 0.7035 \\
CDP-ACC & C3D & \textbf{0.8980} \\ \hline
NLL \cite{cui2020calibrated} & SlowFast & 0.5293\\ 
QR \cite{taylor2000quantile}& SlowFast & 0.4623 \\ 
CQR \cite{romano2019conformalized} & SlowFast & 0.8343 \\
CDP \cite{vovk1999machine}& SlowFast & 0.8140 \\
CDP-ACC & SlowFast & \textbf{0.8407} \\ \hline
$\alpha = 0.1$
\end{tabular}}
\end{table}

\subsubsection{Results of the UQ}
  
Table \ref{baseline} shows the prediction errors of the C3D and SlowFast models under different datasets and training losses. We can find that comparing vanilla regression using MSE or NLL as the learning objective, the prediction error for quantile regression is slightly smaller in AVEC 2013. This indicates the QR with the 50\% quantile has a smaller prediction error than the vanilla regression and NLL. However, in the experiments with the SlowFast model on AVEC 2014, the NLL yields smaller errors compared to the other two methods. This also indicates that if only the prediction accuracy is considered, the NLL method as well as the QR may present better performance. 
  
Table \ref{UQ} shows the results of several uncertainty quantification methods with different depression prediction models. Due to NLL and QR heavily relying on the training process, it's intuitive to observe that the coverage of QR is the worst, followed by NLL. In the current context of depression prediction, unstable training process and imbalanced datasets make it difficult to obtain effective quantile values or fitting Gaussian distributions. For pre-trained QR, CQR is a valid method to calibrate coverage by adjusting the interval width. In this way, CQR can provide adaptive confidence intervals, but its adaptiveness primarily stems from QR, while the conformal prediction mainly provides theoretical guarantees for marginal coverage. In situations where QR exhibits overconfidence, CQR unavoidably relies on average marginal coverage. As for our CDP-ACC, it is an approximate condition coverage method for regression. It can reduce the interval width as much as possible by providing adaptive confidence intervals while maintaining coverage. 
\begin{table}
\centering
\caption{BDI-II scores and depression severity levels}
\label{bdi}{
\begin{tabular}{c c}   
\hline
BDI-II Score & Severity Level \\
\hline
0-13 & minimal \\
14-19 & mild \\
20-28 & moderate\\
29-63 & severe\\
\hline

\end{tabular}}
\end{table}

\begin{figure}
\centering
\includegraphics[width=0.50 \textwidth]{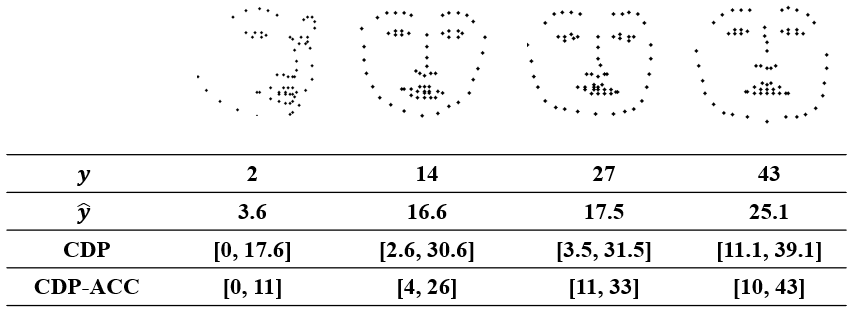}
\caption{The prediction intervals for CDP and CDP-ACC on some sampled video clips with 90\% confidence ($\alpha$ = 0.1). C3D is used as the depression predictor. To protect privacy, we show facial landmarks of the first frame to represent the facial video clips.}
\label{vis}
\end{figure}

In Fig. \ref{vis}, we present several facial video samples with varying levels of depression and their prediction uncertainties, where the predictor uses the C3D regression model. From the figure, we can see that although the predictor exhibits different prediction biases on samples with different levels of depression, CDP and CDP-ACC provide valid confidence intervals for the model's predictions. A wider interval indicates potentially larger uncertainty in the model's predictions. It can also be seen that compared to CDP, CDP-ACC provides tighter confidence intervals while maintaining coverage. On the other hand, we can observe that even though the predictor may have small prediction biases on some samples (e.g., the second example from the left in Fig. \ref{vis}), its prediction uncertainty can still be large. This reminds us that in clinical practice, we cannot focus solely on prediction accuracy. From this perspective, our proposed uncertainty quantification scheme offers new insights and perspectives for assessing potential risks in depression prediction.\newline

\subsubsection{Evaluation of the Conditional Coverage}

Table \ref{SSC} shows the conditional coverage of different UQ methods (with $\alpha = 0.1$) on test instances with various levels of depression severity, which are categorized into four groups ($G=4$, shown in Table \ref{bdi}). A SSC closer to $1 - \alpha$ indicates better conditional coverage of the UQ method. We can clearly observe that NLL and QR perform the worst in terms of SSC due to unstable training and overconfidence. As an effective method for quantifying depression uncertainty, CDP can only guarantee marginal coverage, hence its results are not as good as CQR. However, the conditional coverage of CQR is limited by the pre-trained QR model. In contrast, CDP-ACC can directly approximate conditional coverage based on the point prediction-based vanilla regression models, which does not require retraining a QR model and achieves better conditional coverage.\newline

\begin{figure}
\centering

\includegraphics[width=0.45 \textwidth]{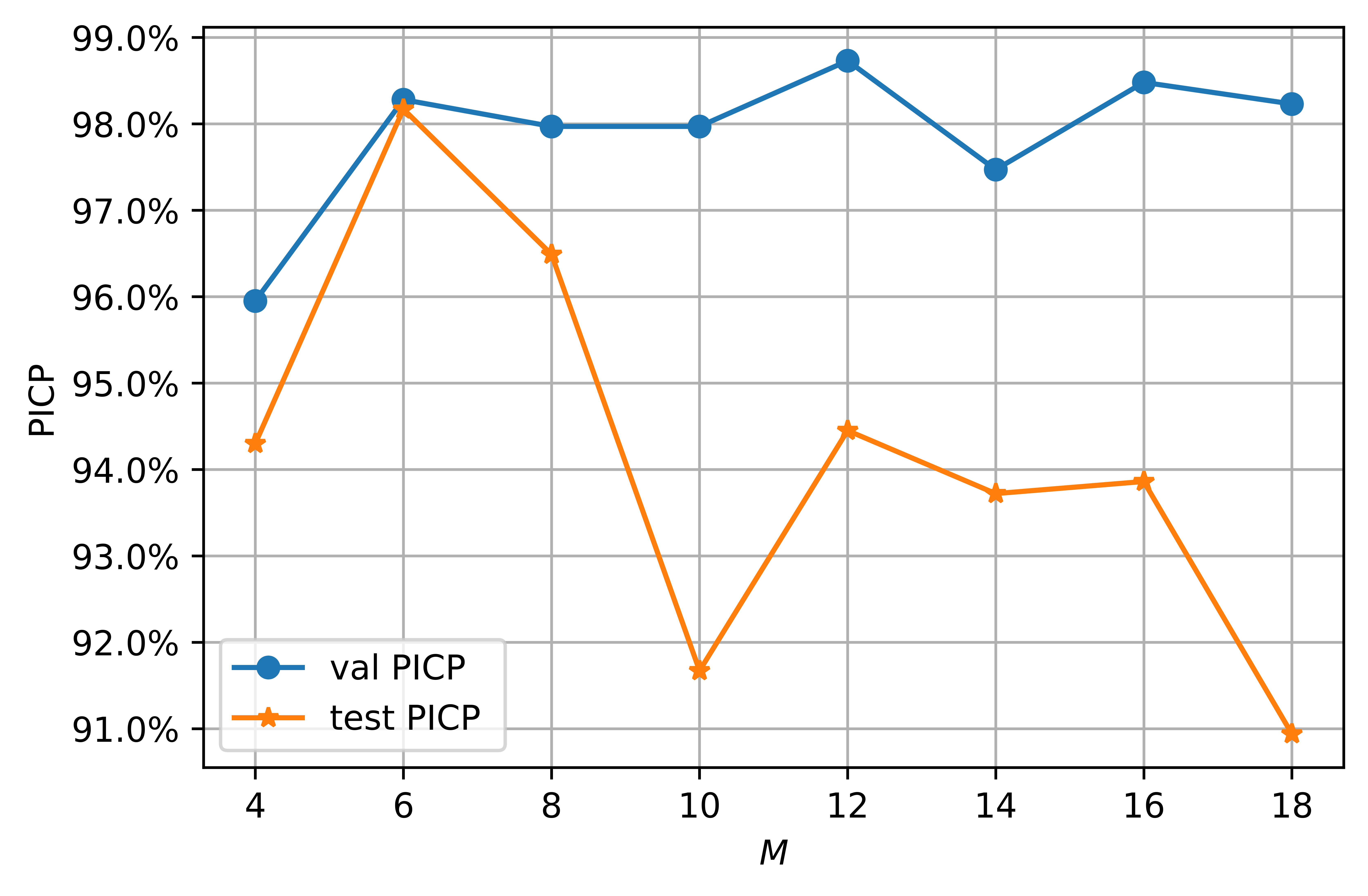}
\caption{PICP of CDP-ACC with varying $M$ on AVEC 2014.}
\label{parm}

\end{figure}
\hfill
\begin{figure}
\hspace{6mm}
\includegraphics[width=0.43 \textwidth]{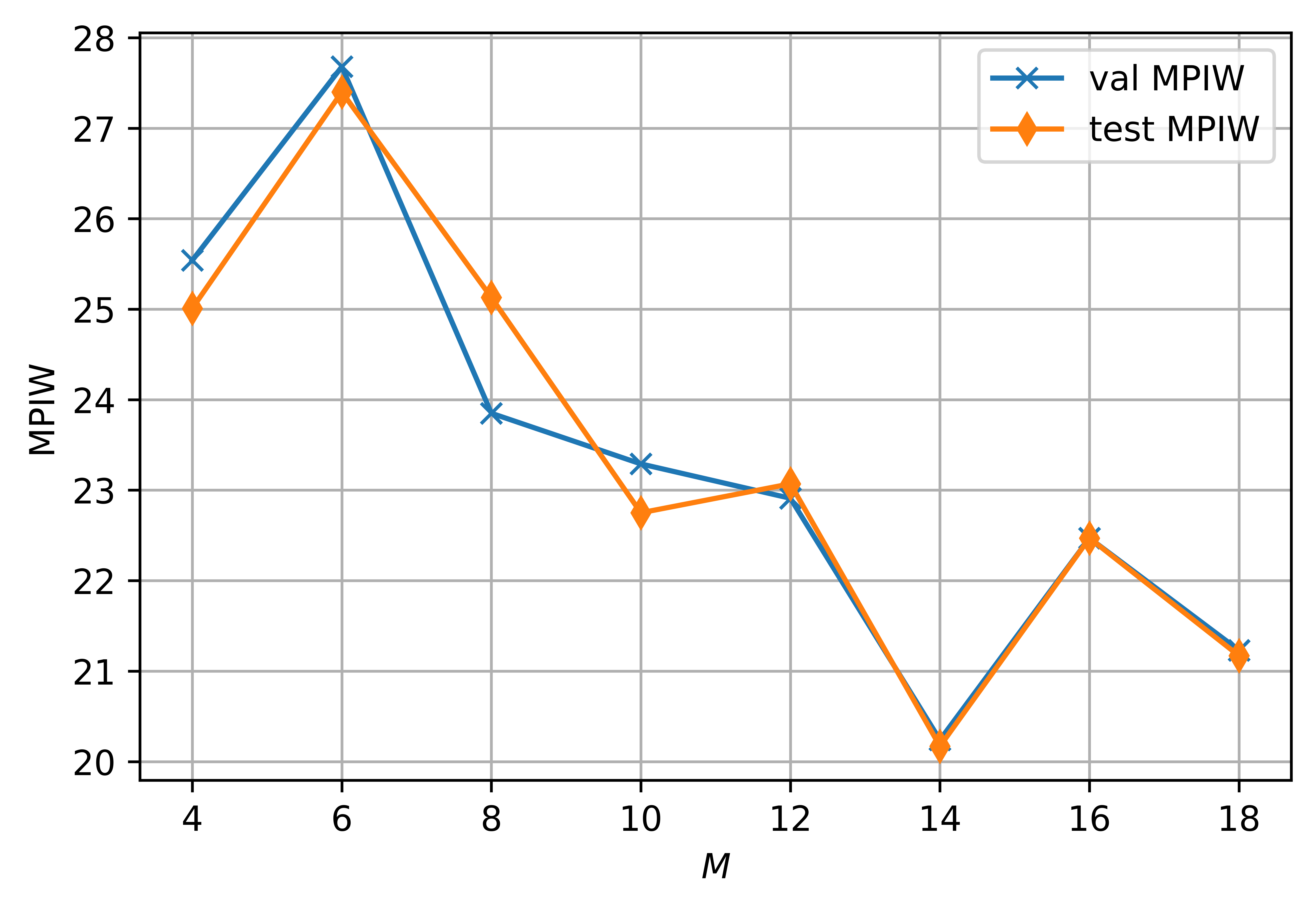}
\caption{MPIW of CDP-ACC with varying $M$ on AVEC 2014.}
\label{parm_pi}

\end{figure}

\subsubsection{Parameter Analysis}

The effectiveness of CDP-ACC largely depends on the accuracy of conditional distribution estimation. In our histogram estimation, the parameter $M$ affects this accuracy. A smaller $M$ implies a coarser histogram estimation, while a larger $M$ results in a finer estimation but reduces the number of sample points within each subinterval $m$, which can also hinder the effectiveness of the approximation. In our experiments, half of the calibration set is used as the validation set for tuning the parameter $M$. As shown in Figs. \ref{parm} and  \ref{parm_pi}, the optimal value of the parameter ($M=14$) can be obtained by tuning on the validation set.

\section{CONCLUSION}
By introducing conformal prediction, we propose a plug-and-play uncertainty quantification method CDP to produce confidence intervals for facial depression prediction. The intervals predicted by CDP can achieve marginal coverage guarantee at any given miscoverage rate. Furthermore, we develop CDP-ACC for facial depression prediction. By approximate conditional coverage, it alleviates the issue of marginal coverage practices focusing solely on the average coverage while overlooking individuals with great uncertainty, and provides valid and adaptive confidence intervals for better uncertainty quantification in facial depression prediction. Although our proposed method is used for facial depression prediction, it is also applicable to various other data modalities. We expect that this work will attract more researchers to pay attention to the issue of uncertainty quantification for automated depression prediction in modern context.


{\appendix
To further validate the effectiveness of our CDP-ACC for general regression tasks, we conducted supplementary experiments on seven publicly available datasets, as also used in \cite{romano2019conformalized,sesia2021conformal}. 
These seven datasets are: \textit{Physicochemical Properties of Protein Tertiary Structure} (bio) \cite{misc_physicochemical_properties_of_protein_tertiary_structure_265}, \textit{BlogFeedback} (blog) \cite{misc_blogfeedback_304}, \textit{Facebook Comment Volume Dataset} \cite{misc_facebook_comment_volume_dataset_363}, variants one (fb1) and two (fb2), from the UCI Machine Learning Repository \cite{asuncion2007uci}, and \textit{Medical Expenditure Panel Survey} (meps19 - meps21) \cite{cohen2009medical}. 
Regarding the preprocessing of the datasets and the experimental setup, we followed the settings in \cite{romano2019conformalized,sesia2021conformal}, with detailed information available on \href{https://github.com/msesia/chr}{GitHub}. 

\subsection*{Datasets}

The \textit{Physicochemical Properties of Protein Tertiary Structure} (bio) dataset \cite{misc_physicochemical_properties_of_protein_tertiary_structure_265} falls under the subject heading of biology. It consists of data extracted from CASP 5-9 and contains 45,730 decoys ranging in size from 0 to 21 angstroms. This dataset is commonly used for multivariate regression analysis.

The \textit{BlogFeedback} (blog) dataset \cite{misc_blogfeedback_304} belongs to the field of social science. It comprises 60,021 instances, each consisting of features extracted from blog posts and a target variable representing the number of comments the post received within 24 hours. This task is a multivariate regression task commonly used to predict the number of comments a post will receive.

The \textit{Facebook Comment Volume Dataset} (fb1, fb2) \cite{misc_facebook_comment_volume_dataset_363} is similar to the blog dataset in that it aims to predict the number of comments on a post. It comprises 40,949 instances, each with 53 features. The dataset includes 5 variants; however, we only utilized the first two variants (fb1 and fb2) in our experiment. 

The \textit{Medical Expenditure Panel Survey} (meps 19 - 21) \cite{cohen2009medical} is a series of large-scale surveys conducted on households, individuals, healthcare providers, and employers across the United States. MEPS serves as the most comprehensive data source on healthcare expenses, utilization, and the extent of medical insurance coverage. In our experiments, we exclusively utilized data from panels 19 to 21.

\subsection*{Implementation Details}

Our experimental setup follows \cite{romano2019conformalized,sesia2021conformal}, where we train a conditional quantile estimator (implemented as a three-layer neural network) to handle the regression task described above. We employ the pinball loss (Eq. \ref{pinball}) as our loss function and ensure that the output quantiles are sorted to prevent crossing. 

We train the neural network using the Adam optimizer with a learning rate of 0.02. The output quantiles serve as part of the post-processing calibration for conformal prediction. Unlike SCP \cite{vovk1999machine,vovk2012conditional,shafer2008tutorial} and CDP-ACC, which do not rely on quantiles, we typically treat the 50\% quantile as the output for vanilla regression. We evaluate using Prediction Interval Coverage Probability (PICP) and Mean Prediction Interval Width (MPIW) as metrics, with a miscoverage rate 
$\alpha$ set to 0.1. For specific definitions of these evaluation metrics, please refer to our main paper.

\subsection*{Results}
We present the uncertainty quantification results of several methods, including SCP \cite{vovk1999machine,vovk2012conditional,shafer2008tutorial}, QR \cite{taylor2000quantile}, CQR \cite{romano2019conformalized}, CHR \cite{sesia2021conformal}, and CDP-ACC, on the aforementioned general datasets. QR and CQR are introduced in the main paper. CHR is a histogram estimation method that calibrates the QR method by constructing nested intervals and integrating them with conformal prediction.

From Table \ref{result}, it is clear that SCP generally achieves marginal coverage but tends to construct unnecessarily conservative intervals. In contrast, in general regression tasks, QR maintains normal coverage rates and effectively reduces the mean interval width (However, in depression prediction task, QR exhibits a trend of overconfidence, with quantile values overly concentrated around predictions, leading to lower coverage rates.) 

CQR and CHR serve as calibration methods for QR, aiming to minimize interval width while maintaining adequate coverage rates. However, CQR is constrained by the simplicity of QR in constructing conformal scores. In contrast, CHR integrates histogram estimation with conformal prediction, enabling more refined optimization of prediction intervals.

For QR, CQR, and CHR, retraining the QR model is crucial. However, vanilla regression models with point prediction can only allow for obtaining prediction intervals through post-processing methods like SCP. Nevertheless, SCP has significant drawbacks such as fixed interval width and overly conservative intervals (unnecessarily wide). CDP-ACC, as an adaptive post-processing method, can be applied to both QR and vanilla regression models. It achieves a finer approximation of conditional distributions through binning, thereby further reducing interval width. From Table \ref{result}, we observe that CDP-ACC achieves competitive uncertainty quantification results, demonstrating its effectiveness on general regression tasks..

\begin{table}
    \centering
    \caption{Results of Uncertainty Quantification.}
		\begin{tabular}{ c c c c }
			\hline
			Methods & Data & PICP & MPIW  \\
			\hline
			SCP \cite{vovk1999machine,vovk2012conditional,shafer2008tutorial} & bio & 90.10\% & 14.87  \\
            QR \cite{taylor2000quantile} & bio & 90.05\% & 14.53  \\
            CQR \cite{romano2019conformalized} & bio & 90.15\% & 14.56  \\
            CHR \cite{sesia2021conformal} & bio & 90.35\% & 13.13  \\
            CDP-ACC & bio & 89.55\% & 12.26  \\
			\hline
			SCP \cite{vovk1999machine,vovk2012conditional,shafer2008tutorial} & blog & 91.20\% & 16.10  \\
            QR \cite{taylor2000quantile} & blog & 92.35\% & 14.65  \\
            CQR \cite{romano2019conformalized} & blog & 91.20\% & 14.65  \\
            CHR \cite{sesia2021conformal} & blog & 89.85\% & 10.24  \\
            CDP-ACC & blog & 90.02\% & 9.56  \\
            \hline
			SCP \cite{vovk1999machine,vovk2012conditional,shafer2008tutorial} & fb1 & 88.25\% & 16.15  \\
            QR \cite{taylor2000quantile} & fb1 & 92.00\% & 15.68  \\
            CQR \cite{romano2019conformalized} & fb1 & 90.00\% & 15.67  \\
            CHR \cite{sesia2021conformal} & fb1 & 89.60\% & 12.23  \\
            CDP-ACC & fb1 & 89.15\% & 12.11  \\
            \hline
			SCP \cite{vovk1999machine,vovk2012conditional,shafer2008tutorial} & fb2 & 90.05\% & 18.51  \\
            QR \cite{taylor2000quantile} & fb2 & 92.55\% & 15.44  \\
            CQR \cite{romano2019conformalized} & fb2 & 89.95\% & 15.43  \\
            CHR \cite{sesia2021conformal} & fb2 & 90.25\% & 12.14  \\
            CDP-ACC & fb2 & 89.70\% & 10.27  \\
            \hline
			SCP \cite{vovk1999machine,vovk2012conditional,shafer2008tutorial} & meps19 & 90.05\% & 32.55  \\
            QR \cite{taylor2000quantile} & meps19 & 89.80\% & 31.23  \\
            CQR \cite{romano2019conformalized} & meps19 & 90.95\% & 31.37  \\
            CHR \cite{sesia2021conformal} & meps19 & 90.04\% & 22.73  \\
            CDP-ACC & meps19 & 89.90\% & 17.00  \\
            \hline
			SCP \cite{vovk1999machine,vovk2012conditional,shafer2008tutorial} & meps20 & 88.85\% & 23.74  \\
            QR \cite{taylor2000quantile} & meps20 & 88.45\% & 26.47  \\
            CQR \cite{romano2019conformalized} & meps20 & 89.10\% & 26.53  \\
            CHR \cite{sesia2021conformal} & meps20 & 89.35\% & 16.56  \\
            CDP-ACC & meps20 & 90.05\% & 16.00  \\
            \hline
			SCP \cite{vovk1999machine,vovk2012conditional,shafer2008tutorial} & meps21 & 90.55\% & 26.58  \\
            QR \cite{taylor2000quantile} & meps21 & 88.80\% & 29.91  \\
            CQR \cite{romano2019conformalized} & meps21 & 90.05\% & 29.99  \\
            CHR \cite{sesia2021conformal} & meps21 & 90.05\% & 19.44  \\
            CDP-ACC & meps21 & 91.15\% & 15.00  \\
            \hline
                
    $^* \alpha = 0.1$ 
    \end{tabular}
    \label{result}
    \end{table}
    }
 
\newpage

\bibliographystyle{IEEEtran}
\footnotesize
\bibliography{ref}


  \begin{IEEEbiography}[{\includegraphics[width=1in,height=1.25in,clip,keepaspectratio]{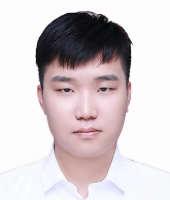}}]{Yonghong Li}
    received a Bachelor of Engineering degree from Chang'an University, Shannxi, China, in 2020.
    He is currently pursuing his PhD degree at Beijing University of Posts and Telecommunications in Beijing, China, with a focus on affective computing and deep learning.
  \end{IEEEbiography}
  

\begin{IEEEbiography}[{\includegraphics[width=1in,height=1.25in,clip,keepaspectratio]{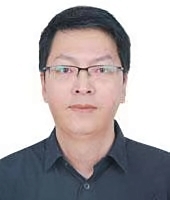}}]{Xiuzhuang Zhou}
  (Member, IEEE) received the PhD degree from the School of Computer Science, Beijing Institute of Technology, China, in 2011. 
  He is currently a full professor at the Beijing University of Posts and Telecommunications, Beijing, China. 
  His research interests include computer vision, pattern recognition, multimedia computing, and machine learning. 
  He has authored more than 80 scientific papers in peer reviewed journals and conferences including several top venues such as the IEEE Transactions on Pattern Analysis and Machine Intelligence, CVPR. 
  He serves as an associate editor for the Neurocomputing.
\end{IEEEbiography}

\end{document}